\documentclass[letterpaper, 10 pt, conference]{IEEEtran}  

\usepackage{geometry}
\usepackage[utf8]{inputenc}
\usepackage{graphicx}
\usepackage{algorithm}
\usepackage[noend]{algpseudocode}
\usepackage{algpseudocode}
\usepackage{cleveref}
\usepackage{enumitem}
\usepackage{booktabs}
\usepackage[font=small]{caption}
\usepackage{cite}
\usepackage{tabularx, colortbl}
\usepackage{multirow}
\usepackage{bm}
\usepackage{siunitx}
\usepackage{verbatim}

\usepackage{balance}
\usepackage{url}

\providecommand{\keywords}[1]{\textbf{\textit{Index Terms---}} #1}

\title{\vspace{0.63cm}\LARGE \bf Demonstration Based Explainable AI for Learning from Demonstration Methods}

\author{Morris Gu,  Elizabeth Croft and Dana Kuli{\'c}\\
Monash University, Australia}

\begin{document}
\maketitle

\begin{abstract}
Learning from Demonstration (LfD) is a powerful type of machine learning that can allow novices to teach and program robots to complete various tasks. However, the learning process for these systems may still be difficult for novices to interpret and understand, making effective teaching challenging. Explainable artificial intelligence (XAI) aims to address this challenge by explaining a system to the user. In this work, we investigate XAI within LfD by implementing an adaptive explanatory feedback system on an inverse reinforcement learning (IRL) algorithm. The feedback is implemented by demonstrating selected learnt trajectories to users. The system adapts to user teaching by categorizing and then selectively sampling trajectories shown to a user, to show a representative sample of both successful and unsuccessful trajectories. The system was evaluated through a user study with 26 participants teaching a robot a navigation task. The results of the user study demonstrated that the proposed explanatory feedback system can improve robot performance, teaching efficiency and user understanding of the robot.  

\end{abstract}  


\section{Introduction}
As robots are introduced into more contexts, including industry~\cite{Goel2020}, residential use~\cite{niemela_telepresence_2021} or service sectors~\cite{service_robotics_2023_2030}, they are being adopted by a greater diversity of users with varied expertise. A powerful paradigm allowing more diverse users to instruct robots is learning from demonstration (LfD)~\cite{ravichandar_lfd_2020}, a process of programming a robot by generalizing from user-provided demonstrations for a given task. LfD is a versatile method which can be applied to various tasks such as grasping~\cite{song_2020_wild}, navigation~\cite{ziebart_maxentirl_2008} and assembly tasks~\cite{Chen2003Programming}. However, novices may be unable to interpret or understand the learnt robot policy due to the black-box nature of learning algorithms and subsequently provide poor teaching. Similar to how human-to-human teaching is improved by feedback~\cite{floden_2017_feedback}, the provision of explanatory feedback can improve human-to-robot teaching~\cite{sena_quantifying_2020, Luebbers2021ARCLFD}. 

Providing explanations about robot or computer behaviour is generally known as \textit{explainable AI} (XAI). XAI researchers hypothesise that such explanations will allow users to better understand and trust these AI systems~\cite{adadi_peeking_2018}. In the context of HRI, the use of XAI can help the user to trust and understand the robot's behaviour~\cite{Silva2023Subjective}. 
However, XAI can also inhibit effective HRI by prompting users to over-rely on provided explanations~\cite{silva_personalized_2024}. Although prior works have explored the impact of XAI on robot performance with LfD~\cite{sena_quantifying_2020, Luebbers2021ARCLFD}, they do not explore how it affects user understanding.
In the LfD context, a lack of user understanding can affect user and robot performance. Poor understanding could lead users to provide suboptimal demonstrations, which may negatively impact both learning performance and efficiency~\cite{sena_quantifying_2020}. 

\begin{figure}[t]
    \centering
    \includegraphics[trim = {0 0 0 0}, clip, width=0.9\linewidth]{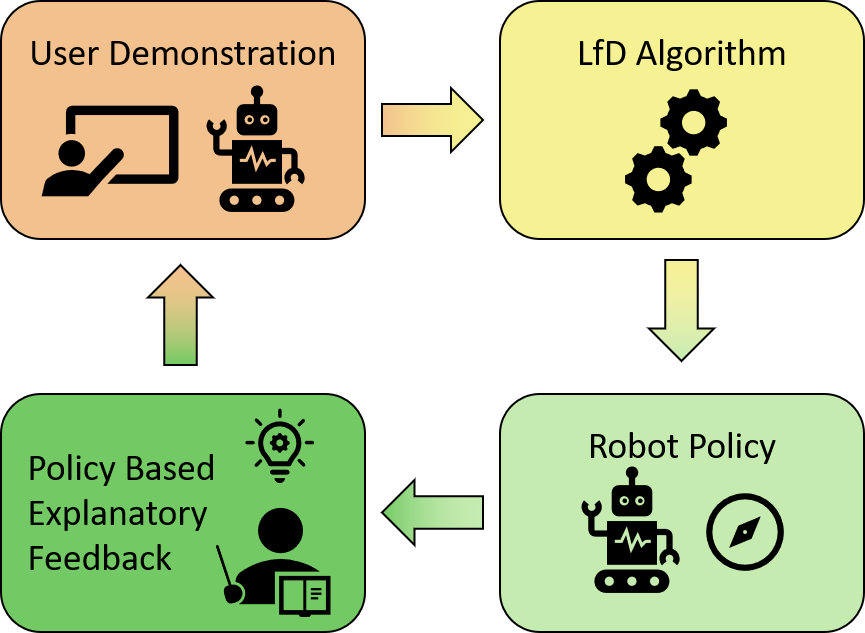}
    \caption{High-level overview of a learning from demonstration system that incorporates policy explanatory feedback. The explanatory feedback module is designed to improve human-to-robot teaching through greater user understanding of learnt robot behaviour}
    \vspace{-0.4cm}
    \label{fig:method}
\end{figure} 

This work proposes an adaptive-sampling and demonstration-based explanation system which can automatically generate explanations of the robot's learned policy. We apply this system to a navigation teaching task where users teach a robot by demonstrating trajectories. After the robot learns a policy based on a user's demonstrated trajectories, we provide selected samples of the robot's successful and unsuccessful task execution using the learnt policy - showing the user what the robot can and cannot do. We hypothesise that this feedback will improve user understanding of the robot's policy, leading to better demonstrations that result in faster robot learning.  A user study was conducted with this system, evaluating established metrics of robot performance~\cite{sena_quantifying_2020, sakr2023everyday} and user certainty and understanding.
The paper makes the following contributions:

\begin{enumerate}[leftmargin = 15pt]
    \item We implement a novel demonstration-based explanatory feedback system which is agnostic the LfD algorithm used  and adapt to the user's teaching
    \item The first work, to the authors' best knowledge, that evaluates whether XAI can improve a user's understanding when applied to an LfD algorithm
    \item We show that explanatory feedback improves user teaching and understanding, and ultimately learnt robot performance within LfD
\end{enumerate}

\section{Related Work}
\label{sec:related} 
XAI is the concept of providing explanations or interpretations of AI systems to users. It is an important aspect to consider when designing and implementing machine learning and/or AI systems~\cite{adadi_peeking_2018, barredo_arrieta_2020_XAI}. Within HRI, interactions between users and robots can benefit from provided explanations~\cite{sena_quantifying_2020, paleja_2021_utility, Silva2023Subjective, silva_personalized_2024}. More specifically, XAI-supplemented LfD is an effective method to improve robot performance following human-to-robot teaching~\cite{sena_quantifying_2020}.

\textbf{Non-HRI XAI Approaches}: XAI has been used to provide explanations regarding various types of AI systems~\cite{gunning_XAIexplainable_2019}. In early works, XAI often supplemented image classification networks, through either intrinsic or post-hoc methods. Intrinsic methods implement XAI directly within the AI system whereas post-hoc methods involve using external tools such as supplementary networks to provide explanations. XAI in a classification network can be implemented by providing implicit or explicit semantic information such as highlighting the importance of regions or pixels within an image~\cite{Bach2015Pixelwise, ribeiro-etal-2016-lime,Ribeiro_anchors_2018, qiang_graphlime_2022}, associating inputs with approximated rules~\cite{ribeiro-etal-2016-lime,Ribeiro_anchors_2018} or simultaneously generating natural language descriptions of an image~\cite{xu_2015_show_attend}. Although intrinsic methods such as the work from Bach et al.~\cite{Bach2015Pixelwise} and Xu et al.~\cite{xu_2015_show_attend} can allow a traditionally black box system to become explainable/interpretable, this can become difficult when models become more complex~\cite{adadi_peeking_2018}. In these situations, post-hoc methods such as surrogate models~\cite{ribeiro-etal-2016-lime,Ribeiro_anchors_2018, qiang_graphlime_2022} can be useful. Post-hoc methods can be used to explain any black-box system which has the same task, although this can require a different and interpretable representation from the original model/algorithm~\cite{ribeiro-etal-2016-lime,Ribeiro_anchors_2018}. However, this extra effort allows the method to be agnostic to the machine learning system or model, allowing it to be both versatile and generalizable. This is particularly useful for our work, where we apply a post-hoc method, aiming to generalize to any LfD algorithm. 

\textbf{XAI and HRI}: XAI methods in the context of HRI provide explanations of actions or decisions taken by the robot that relate to an interaction or shared task. These explanations can be provided in the form of natural language~\cite{das_explainable_2021, Silva2023Subjective, silva_personalized_2024}, trajectory demonstration~\cite{sena_quantifying_2020, adamson_why_2021}, trajectory visualization~\cite{sena_quantifying_2020, Mueller2021Counterfactuals, Luebbers2021ARCLFD}, or decision trees~\cite{paleja_2021_utility, Silva2023Subjective}. Though different XAI methods have been shown to similarly improve human-robot/agent interactions~\cite{Silva2023Subjective}, implementing XAI can involve trade-offs such as an overreliance on provided explanation~\cite{silva_personalized_2024} and decreased task performance when considering intrinsic methods~\cite{adadi_peeking_2018}. Another potential trade-off of XAI includes an increased human mental load, which can lead to an unimproved or even decreased performance in a human-robot collaboration task~\cite{paleja_2021_utility}. As such it is important to effectively consider a user's mental load.

\textbf{XAI and LfD}: Effectively integrating XAI into LfD is crucial for improving human-to-robot teaching, particularly for novice users. 
Luebbers et al.~\cite{Mueller2021Counterfactuals} used counterfactuals, the process of negating certain causal conditions, to explain what the robot has learnt. This was done by visualizing the robot's trajectory with and without a certain learnt parameter in augmented reality. This allowed users to intuit the causal relationship between their teaching and the robot's learnt behaviour. Sena et al.~\cite{sena_quantifying_2020} studied the effects of XAI on LfD by visualising to the user trajectories generated by the learned policy from different locations in the workspace. They demonstrated that it improved human teaching and subsequent robot performance, but only with explanations that conveyed the level of generalization of the learned policy. However, explanations from already taught or user-chosen locations did not significantly improve teaching performance.
They observed this through two user studies, a 2D point-to-point reaching and a pick-and-place task. The 2D point-to-point user study also demonstrated that feedback removed the need for explicit rule-type guidance. However, a key limitation for current research on XAI and LfD~\cite{sena_quantifying_2020, Luebbers2021ARCLFD} is their reliance on hand-crafted explanations which become less scalable as a robot's potential state space increases. Our work aims to address this issue by proposing an adaptive sampling system which can generate trajectories automatically regardless of state space size. The proposed system adapts to the user's teaching by selectively demonstrating task executions that are successful and unsuccessful.

\textbf{Evaluation Metrics}: It can be challenging to effectively quantify the benefits and limitations of XAI on LfD. Sena et al.~\cite{sena_quantifying_2020}, introduce a set of metrics to measure teaching efficiency and efficacy. They define efficacy as the proportion of successful task realizations to all the possible ways of performing the task successfully. Efficiency is then obtained by dividing efficacy by the number of provided user demonstrations. These metrics are additionally used by Sakr et al.~\cite{sakr2023everyday} to quantify LfD. However, human-centred metrics are missing, such as user understanding and/or trust, that have been explored in computational/virtual agents~\cite{Silva2023Subjective, Weitz2021Explain}. For example, Weitz et al. found that the integration of virtual agents and XAI improved user trust~\cite{Weitz2021Explain}. Additionally, Silva et al. investigated the effects of a wide range of approaches, finding that XAI, in general, is correlated with trust, however, the choice of XAI approach can result in differences in understanding~\cite{Silva2023Subjective}. Given the impact of XAI on user understanding and trust, the risks of XAI overreliance~\cite{silva_personalized_2024} and a potential increased mental load~\cite{paleja_2021_utility}, it is crucial to accurately measure these factors. This is particularly important in the case of user understanding, as it has not been explored in works integrating XAI and LfD. To address this, we evaluate user perceptions of our proposed system, such as mental load, and develop a method to quantify user understanding and certainty within the context of LfD. 


\begin{figure}[t]
    \centering
    \includegraphics[trim = {0 0 0 0}, clip, width=1\linewidth]{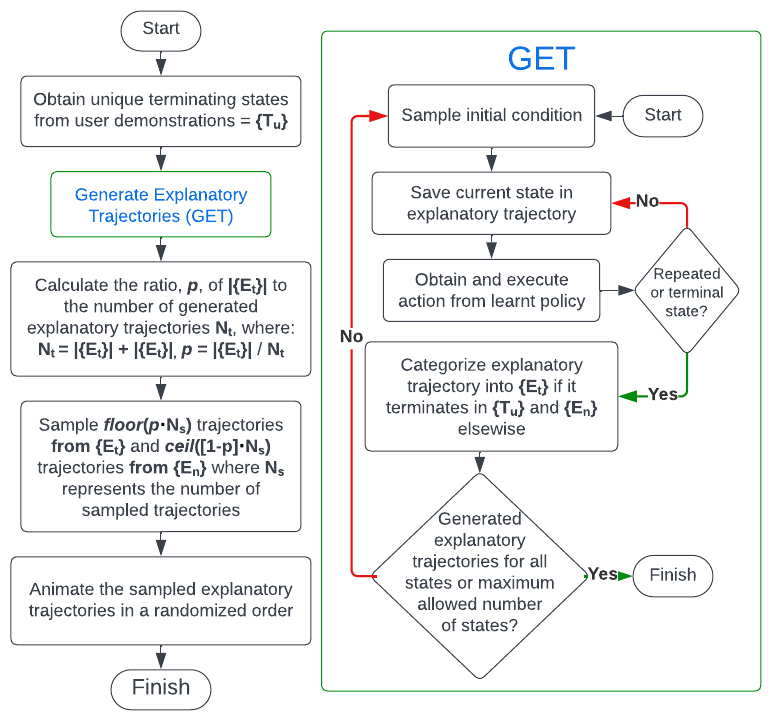}
    \vspace{-0.13cm}
    \caption{A flow diagram for generating and visualizing explanatory trajectories including the specific process for generating and categorizing trajectories (GET), highlighted in the green box to the right.} 
    \vspace{-0.7cm}
    \label{fig:explanation_generation}
\end{figure}

\section{Adaptive Explanatory Feedback for LfD}
\label{sec:method}
In this work, we present an adaptive-sampling explanatory feedback system for LfD algorithms. The proposed system is implemented into a multi-goal navigation task, where the user's objective is to teach a robot to navigate to a set of goals. Feedback is implemented through \textit{explanatory trajectories}, similar to Sena et al.~\cite{sena_quantifying_2020}. These are categorized as successful or unsuccessful based on the given task, with \textit{explanatory trajectories} generated to effectively cover the state space. \textit{Explanatory trajectories} are then selectively sampled to maintain the ratio between the successful and unsuccessful trajectories. We sequence user demonstrations, robot learning and explanations in an iterative manner, as shown in Figure~\ref{fig:method}. This creates a positive feedback loop, enabling the user to better understand the robot's current capability, improving subsequent demonstrations and learnt robot policies. 

\subsection{Method}
\label{subsec:gen_exp}

\begin{figure}[t]
    \centering
    \includegraphics[trim = {0 0 0 0}, clip, width=0.75\linewidth]{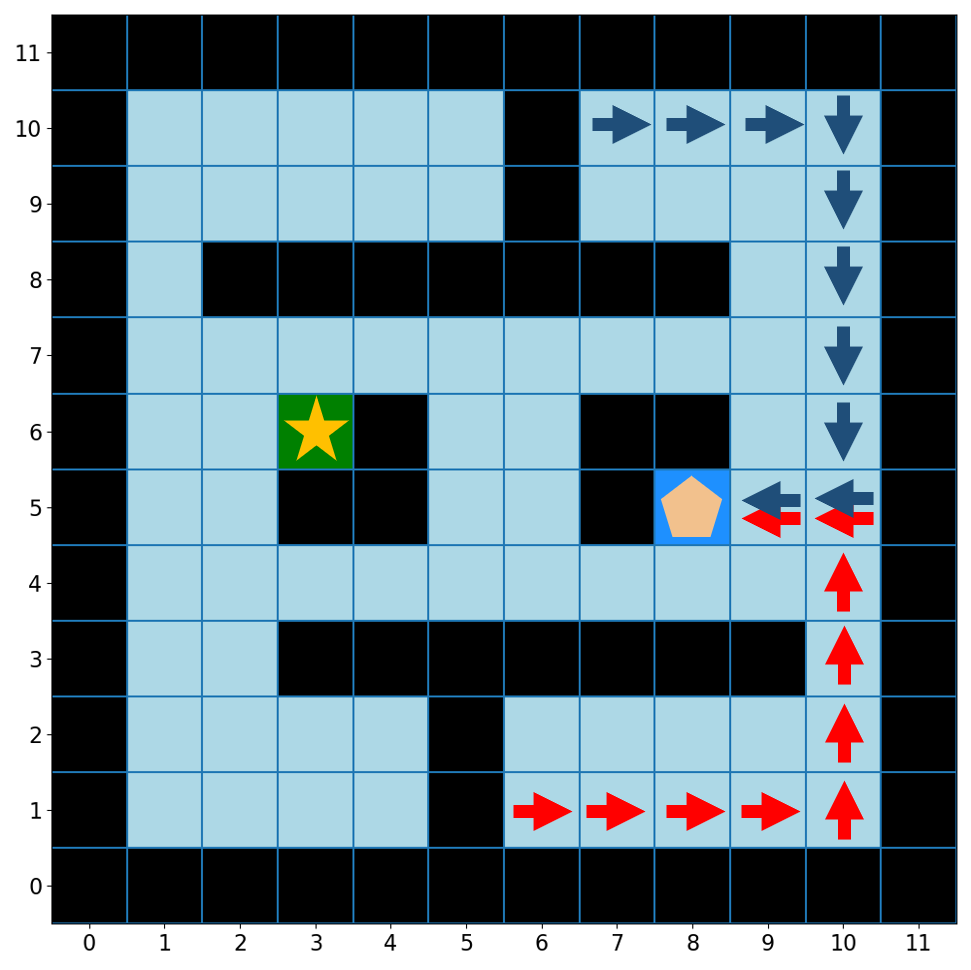}
    \vspace{-0.2cm}
    \caption{An example of the grid world environment employed in the user studies. Example demonstrations are visualized as red and blue arrows, originating from the top and bottom of the grid world respectively. In the diagram, the blue goal, indicated by the pentagon, is considered a \textit{user-demonstrated goal} as the terminating state of both demonstrations is this goal. The green goal, indicated by a star, is not a \textit{user-demonstrated goal}, as no demonstrations reach this goal.}
    \vspace{-0.5cm}
    \label{fig:grid_world}
\end{figure}

This work proposes an adaptive-sampling and explanatory feedback system to explain the output of an LfD algorithm. Explanatory trajectories are generated by executing the learnt LfD policy derived from user-provided demonstrations. The policy generates successive actions, and simulates their effects in a noise-free environment, to generate either successful or unsuccessful trajectories based on the set of terminating states found within user-demonstrations, ${T_u}$. We define a successful \emph{explanatory trajectory} as a trajectory that terminates in a state contained within this set. This definition is visualized in Figure~\ref{fig:grid_world}.
Considering this definition, the system generates and categorizes a population of explanatory trajectories. A subset of these trajectories is then adaptively sampled such that the ratio between successful and unsuccessful \textit{explanatory trajectories} of the population of trajectories is maintained within the subset of sampled trajectories, rounding up the number of unsuccessful \textit{explanatory trajectories} when the ratio does not produce an 
exact integer amount of each category. This subset of trajectories is then animated for users. The process for generating explanatory trajectories is illustrated in the flow diagram in Figure~\ref{fig:explanation_generation}. 

\subsection{Implementation Details} 
The system was implemented on the grid world environment shown in Figure~\ref{fig:grid_world}. During user demonstrations, there was a fixed 20\% chance for the robot to move perpendicular to the user's input action, serving to generate noisy demonstrations. A reward function is learnt by employing MaxEntIRL~\cite{ziebart_maxentirl_2008} on the user-provided demonstrations. The features were state-based, with each state being represented by one-hot encoded vectors. The robot policy is then derived from the reward function by applying value iteration to it. The value iteration used a discount factor of $\gamma = 0.9$ and assumed an infinite horizon, with the robot remaining stationary after the goal state is reached. The overall process for collecting demonstrations and learning the robot policy was as follows:

\begin{figure*}[t]
    \centering
    \includegraphics[trim = {0 38 0 38}, clip, width=0.95\linewidth]{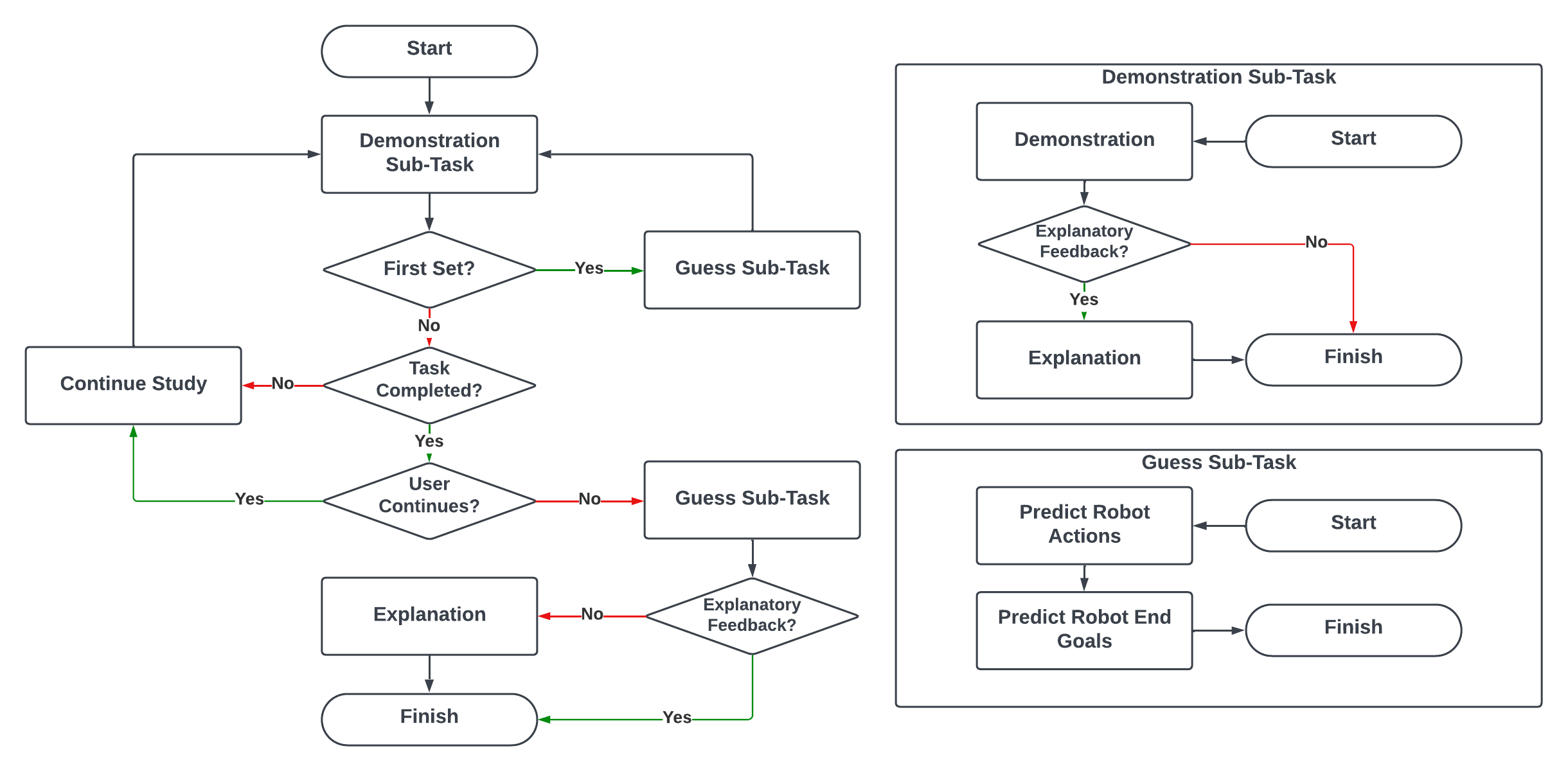}
    \vspace{-0.15cm}
    \caption{A flow diagram for the user study which indicates the order of user study sub-tasks within the teaching navigation task.} 
    \vspace{-0.5cm}
    \label{fig:study_structure}
\end{figure*}

\begin{enumerate}
    \item Collect trajectories from a participant in a given grid world.
    \item Infer a reward function from these trajectories with MaxEntIRL~\cite{ziebart_maxentirl_2008}, assuming state-based features.
    \item Apply value iteration on the learnt reward to generate a policy. 
\end{enumerate}
\vspace{-0.3cm}

\section{User Study}
\label{subsec:inituserstudytask}
We validated this approach by conducting an in-person user study. This study was approved by the Monash Human Research Ethics Committee (Project ID: 35674) 

\subsection{Navigation Task Details}
Participants in the user study completed a multi-goal navigation task on a grid world environment shown in Figure~\ref{fig:grid_world}. The user's objective was to teach the robot how to navigate to a set of goals from any state within the environment. The overall task consisted of the following sub-tasks: 
\begin{enumerate}
    \item \textbf{Demonstration} - The participant provides 5 valid teaching demonstrations of the task learnt by the robot by navigating it in a \textit{stochastic} grid world environment from a participant-chosen starting position to a goal state. A valid teaching demonstration is any demonstration which reaches a goal state. During this demonstration sub-task, the participant is informed of a preferred goal location and a non-preferred goal location, and a variable action budget based on the robot's distance to the preferred goal. Given the action budget, participants were instructed to demonstrate trajectories to a preferred goal if reachable within the budget, or the non-preferred goal otherwise.
       
    \item \textbf{Explanation} - The participant views 5 \emph{explanatory} trajectories as described in Section~\ref{subsec:gen_exp}.
    
    \item \textbf{Action Prediction} - The participant predicts the robot's learnt action from 5 random states and rates the certainty of each prediction as a Likert scale variable (1-7).
    
    \item \textbf{Goal Prediction} - The participant is presented with 5 random initial states and is asked to predict which goal the robot will reach after generating successive learnt policy actions from the states in a noise-free environment. The participant can choose to predict if the robot will terminate in the preferred goal, non-preferred goal or no goal. They rate the certainty of each prediction as a Likert scale variable (1-7).
\end{enumerate}
The goal and action \emph{prediction} sub-tasks were implemented to respectively measure a user's high- and low-level understanding of the robot's behaviour. 

\subsection{User Study Structure and Details}
\label{subsec:structure}
The study consisted of two conditions, \emph{explanatory feedback} (EF) and \emph{no feedback} (NF). Each participant teaches an agent to perform a grid world navigation task under one of the conditions (between-participant design). The study structure is shown in Figure~\ref{fig:study_structure}. The only difference between conditions is that one group (EF) receives explanations as feedback, whilst the other group does not get feedback (NF) (although they see an explanation at the end of the study). For the EF condition, participants view \textit{explanatory trajectories} after each \emph{demonstration} set. Participants in the NF condition instead only viewed \textit{explanatory trajectories} after the final set of predictions. The purpose of post-task explanations is to allow participants to view the robot's behavior without impacting their teaching. Additionally, participants completed two \textit{action prediction} and two \textit{goal prediction} sub-tasks after the initial and final set of provided demonstrations.

During the user study task, the user is additionally shown a \emph{robot performance} value, which is the percentage of states which generate successful \textit{explanatory trajectories} as defined in Section~\ref{subsec:gen_exp}. This \emph{performance} value is calculated after each \textbf{demonstration} sub-task (i.e., 5 valid teaching demonstrations). The task is considered completed when the participant reaches greater than 95\% performance, or after completing 10 \textbf{demonstrations} sub-tasks. After this, participants were allowed to voluntarily continue or finish the user study.

\subsection{Experimental Protocol}
\label{subsec:expPro}
The study was conducted in a private meeting room and performed on a laptop computer set up for the study. During the studies, the participants first read the explanatory statement and filled in a consent form and pre-study demographic survey. Following this, they watched an explanatory video for the task and then participated in a practice session to become acquainted with the task and system dynamics. During this practice session explanatory feedback was generated with a pre-defined policy. The participant was informed that the provided feedback in this practice session was not generated from their own teaching.
Throughout the user study task, the experimenter read from a script to maintain uniformity of procedure. After completing the task, the participant filled in a post-task survey with questions from Table~\ref{tab:questions} and was invited to provide optional general feedback in a text box and more specific feedback in an interview. The duration of the user study was between 20 and 30 minutes. 

\subsection{Participants and Metrics}
\label{subsec:objmet}
We recruited 26 participants from Monash University staff and students on campus. In a demographic survey, participants rated their experience in robotics and machine learning (ML) from 1-7. 85\% and 76\% of participants rated themselves  as 4+ (Robotics: $\mu = 4.31, \sigma = 1.09$, ML: $\mu = 4.27, \sigma = 1.46$).
\\

\textbf{Prediction Metrics}:
The following metrics measure the goal and action prediction sub-tasks. 
\begin{itemize}
    \item \textbf{Prediction Accuracy}: We measured \emph{action prediction accuracy} and \emph{goal prediction accuracy} as the number of correct action and goal predictions in the prediction sub-tasks. The \emph{initial} and \emph{final} sub-task instances were each measured separately for each type of prediction sub-task.
    \item \textbf{Prediction Certainty}: We measured \emph{action prediction certainty} and \emph{goal prediction certainty} in the prediction sub-task on a 7-point Likert scale for each prediction. The \emph{initial} and \emph{final} sub-task instances were each measured separately for each type of prediction sub-task.
    \item \textbf{Prediction Time}: We measured \emph{action prediction time} and \emph{goal prediction time} by summing all prediction sub-task times. The times were measured from when a sub-task window was opened until when it was closed. 
\end{itemize}

\textbf{Performance Metrics}: 
The performance metrics are adapted from Sena et al.~\cite{sena_quantifying_2020} and Sakr et al.~\cite{sakr2023everyday}. 
\begin{itemize}
    \item \textbf{Robot Performance}: The percentage of states from which a goal state is reached in a noise-free environment when following the policy learned from user demonstrations. 
    \item \textbf{Number of Demonstrations}: The number of demonstrations provided by the participant to teach the robot.   
    \item \textbf{Teaching Efficiency}: \emph{Robot Performance} divided by the \emph{Number of Demonstrations}. 
    \item \textbf{Early State Teaching Efficiency}: \emph{Teaching efficiency} after four demonstration sub-tasks (20 demonstrations). 
    This metric was designed to capture early teaching efficiency as we observed in pilot studies that teaching efficiency was higher for initial sets of observations. Measurement after four demonstration sub-tasks was selected as this was designed as the earliest point at which participants could complete the task.
\end{itemize}

\textbf{Perception Metrics}:
\renewcommand{\arraystretch}{1.1}
The questions in Table~\ref{tab:questions} were asked to gauge participant perceptions. Questions marked with an "R" indicate that question has a reverse scale.
\begin{table}[h]
\centering
\vspace{-0.1cm}
\caption{Study Survey Questions)}
\begin{tabular}{|l|}
\hline 
    Q1: I feel satisfied with the robot’s behaviour. (\textbf{Robot Satisfaction})\\ 
    Q2: I understand what the robot learnt. (\textbf{Understanding Perception})\\
    Q3: How would you rate your teaching? (\textbf{Teaching Perception})\\
    Q4: How mentally demanding was the task? (\textbf{Mental Demand}) (R)\\
\hline
\end{tabular}
\vspace{-0.3cm}
\label{tab:questions}
\end{table}
\renewcommand{\arraystretch}{1}

\subsection{Hypotheses}
\label{subsec:hypotheses}
For this study, we formulated the following hypotheses to compare the two conditions - \emph{explanatory feedback} (EF) and \emph{no feedback} (NF).
\begin{enumerate}[label=\textbf{H\arabic*}]
    \item Providing feedback during the task will improve the learnt policy of the robot (robot performance).
    \item Providing feedback during the task will improve teaching efficiency.
    \item Providing feedback during the task will improve participant understanding of the robot's learnt behaviour by improving the goal and action prediction accuracy.
    \item Providing feedback during the task will improve participant understanding certainty through improved prediction certainty.
    \item Providing feedback during the task will improve participant perception by increasing: (a) satisfaction with the robot, (b) perceived understanding , (c) perceived teaching performance and (d) reducing mental demand.
\end{enumerate}

\section{Results}
\label{sec:results}

\subsection{Data Analysis}
\begin{figure}[ht]
\centering
        \includegraphics[trim = {0 10 0 0}, clip, width=0.85\linewidth]{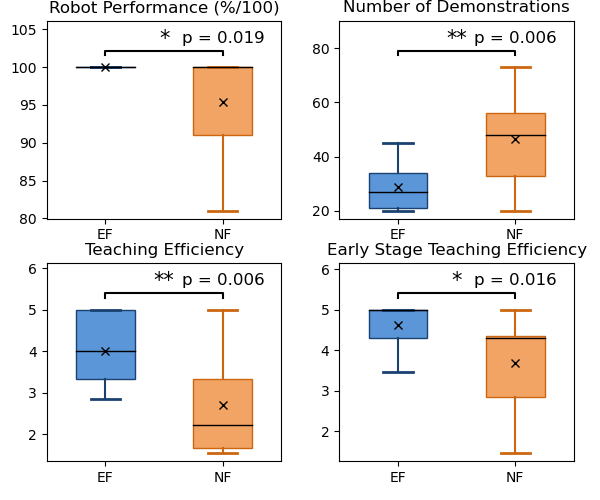}
        \vspace{-0.2cm}
        \caption{Distributions of the Performance metrics, relating to \textbf{H1} (Performance) and \textbf{H2} (Teaching Efficiency), across the \emph{explanatory feedback} (EF) and \emph{no feedback} (NF) conditions. Means are indicated by crosses, with the level of statistical significance indicated by asterisks adjacent to the corresponding p-value (0 $<$ *** $<$ 0.001 $<$ ** $<$ 0.01 $<$ * $<$ 0.05 $<$ ns)}
        \label{fig:performance_bar}
\end{figure}

\begin{figure}[ht]
\centering
        \includegraphics[trim = {0 10 0 0}, clip, width=0.88\linewidth]{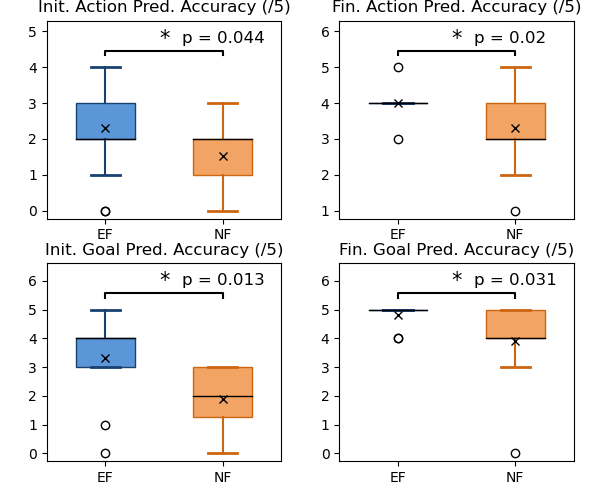}
        \caption{Distributions for \textbf{H3} metrics, goal and action prediction accuracy, across the \emph{explanatory feedback} (EF) and \emph{no feedback} (NF) conditions. Means are indicated by crosses, outliers are indicated with open circles, with the level of statistical significance indicated by asterisks adjacent to the corresponding p-value (0 $<$ *** $<$ 0.001 $<$ ** $<$ 0.01 $<$ * $<$ 0.05 $<$ ns)}
        \vspace{-0.4cm}
        \label{fig:accuracy_bar}
\end{figure}

\begin{figure}[ht]
\centering
        \includegraphics[trim = {0 10 0 0}, clip, width=0.88\linewidth]{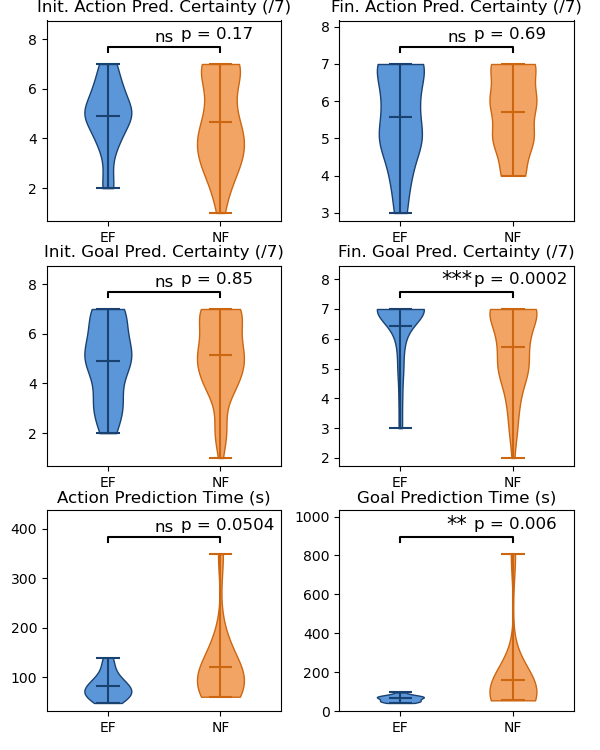}
        \caption{Distributions for prediction certainties, relating to \textbf{H4}, and prediction times, across the \emph{explanatory feedback} (EF) and \emph{n feedback} (NF) conditions. The level of statistical significance is indicated by asterisks adjacent to the corresponding p-value (0 $<$ *** $<$ 0.001 $<$ ** $<$ 0.01 $<$ * $<$ 0.05 $<$ ns)}
        \vspace{-0.3cm}
        \label{fig:prediction_bar}
\end{figure}

\begin{figure}[ht]
\centering
        \includegraphics[trim = {0 10 0 0}, clip, width=0.88\linewidth]{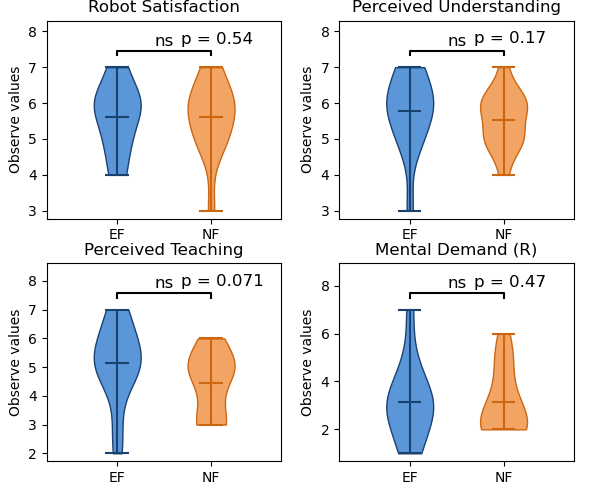}
        \caption{Distribution of responses to Survey questions on user perception, relating to \textbf{H5}, comparing the \textbf{explanatory feedback} (EF) and \textbf{n feedback} (NF) conditions. Greater values indicate better perception except Mental Demand which has a reverse scale}
        \vspace{-0.4cm}
        \label{fig:barplot}
\end{figure}

We first tested for normality with the Shapiro-Wilk test but could not confirm normality for every variable. As none of our variables met the null-hypothesis of a normal distribution, we visualized all variables on a Q-Q plot and determined that all variables were not normally distributed. We applied the Mann-Whitney U test on all of our data; results are summarized in Figures~\ref{fig:performance_bar}, \ref{fig:accuracy_bar}, \ref{fig:prediction_bar} and \ref{fig:barplot}. We note that the data for \emph{goal predictions} excluded six participants due to a data processing error, although the remaining data was still considered valid as they were not made aware of the error.

\begin{figure}[h]
\vspace{-0.4cm}
\centering
        \includegraphics[trim = {10 20 10 10}, clip, width=0.85\linewidth]{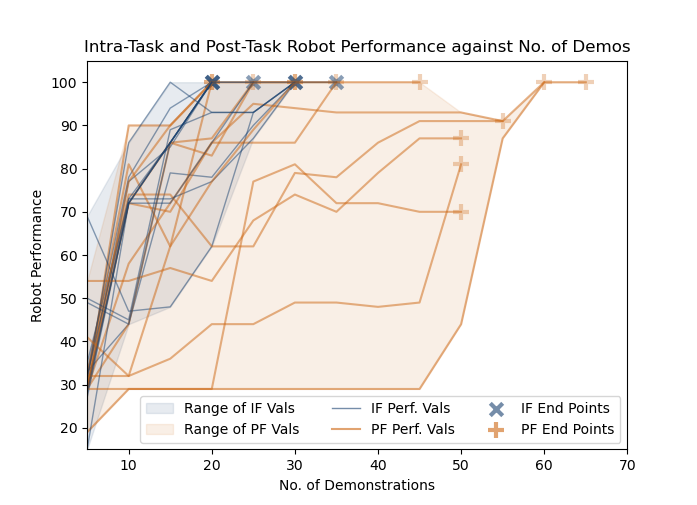}
        \vspace{-0.2cm}
        \caption{A graph plotting each user's performance against the number of demonstrations. \emph{explanatory feedback} (EF) condition participants are visualized in blue; \emph{no feedback} (NF) condition participants are visualized in orange. Crosses and Plusses indicate when each user ended/completed the user study task and the shaded regions indicate the range of values for each condition.}
        \vspace{-0.4cm}
        \label{fig:performance}
\end{figure}

\subsection{Performance Metrics}
Robot performance (\textbf{H1}) and teaching efficiency (\textbf{H2}) were analyzed by comparing performance metrics. For \textbf{H2}, we compared the \emph{number of demonstrations}, \emph{teaching efficiency} and \emph{early stage teaching efficiency}. We found significant improvement in all these performance metrics, as can be seen in Figure~\ref{fig:performance_bar}, affirming both \textbf{H1} and \textbf{H2}. Analyzing the results for robot performance (\textbf{H1}) we note that all participants in the EF condition achieved 100\% robot performance, unlike those in the NF condition. Figure~\ref{fig:performance} shows the time evolution of the robot performance for each of the participants. Six participants in the NF condition ended the study after 10 demonstration sets and 4 of them were unable to achieve 100\% performance. Furthermore, the greater teaching inefficiency in the NF condition is shown inFigure~\ref{fig:performance} largely due to participant performance plateaus after varying numbers of trajectories.

\begin{table}[h]
    \centering
    \caption{Summarized results from the Wilcoxon Sign-rank test comparing \emph{initial} and \emph{final} prediction accuracy for the goal and action prediction sub-tasks within the \emph{explanatory feedback} (EF) and \emph{no feedback} (NF) conditions. \emph{Initial} and \emph{final} sub-task results are indicated by \textit{I} and \textit{F} respectively. The results demonstrate that prediction accuracy improves after teaching. The level of statistical significance is indicated by asterisks adjacent to the corresponding p-value (0 $<$ *** $<$ 0.001 $<$ ** $<$ 0.01 $<$ * $<$ 0.05 $<$ ns)}
   
    \begin{tabular}{|c|c|c|c|c|}
    \hline
         Metric & $\bm{p}$ & $\mu_I + \sigma_I$ & $\mu_F + \sigma_F$ & \(U_F\) \\\hline
         NF Action Pred. & \textbf{0.001}** & 1.54 + 0.75 & 3.31 + 1.07 & 148 \\
         EF Action Pred. & \textbf{0.002}*** & 2.31 + 1.32 & 4.00 + 0.39  & 152.5\\
         NF Goal Pred. & \textbf{0.005}** & 1.90 + 1.14 & 3.9 + 1.45 & 84.0 \\
         EF Goal Pred. & \textbf{0.008}** & 3.30 + 1.55 & 4.80 + 0.40 & 89.0 \\\hline
    \end{tabular}    
    \label{tab:init_vs_fin_preds}
    \vspace{-0.4cm}
\end{table}

\subsection{Prediction and Perception Metrics}
We tested whether participant understanding (\textbf{H3}) and participant prediction certainty (\textbf{H4}) were improved by providing feedback by comparing prediction accuracy and certainty metrics respectively. In Figure~\ref{fig:accuracy_bar}, each prediction accuracy metric was significantly improved by providing feedback during the task, affirming \textbf{H3}. Despite this, Figure~\ref{fig:prediction_bar} shows that users did not demonstrate a higher certainty overall so we cannot affirm \textbf{H4}, although \emph{final goal prediction certainty} was significantly higher. However, although participants were not more certain of their predictions, they still appeared to be quicker at predicting when given feedback during the task, with \emph{goal prediction time} being significantly improved by explanatory feedback while \emph{action prediction time}, though borderline not significant, did have a lower mean and variance. A Wilcoxon Sign Ranked Test was additionally applied to prediction accuracy across the initial and final sub-tasks. The results, summarized in Table~\ref{tab:init_vs_fin_preds}, show that prediction accuracy is significantly improved after completing the teaching task, compared to initial prediction accuracy, regardless of condition.

To determine whether the user's perception was improved by explanatory feedback \textbf{(H5)}, we compared the responses to the survey questions in Table~\ref{tab:questions}, with summarized data shown in Figure~\ref{fig:barplot}. Overall, none of the metrics showed any significance supporting \textbf{H5}. Though we cannot affirm \textbf{H5}, the visual similarities in Figure~\ref{fig:barplot} suggest that providing explanatory feedback results in similar perceptions whilst providing benefits for robot performance, teaching efficiency and user understanding.

\section{Discussion}
\label{subsec:disc}

\textbf{Performance}: Consistent with prior research~\cite{sena_quantifying_2020}, explanatory feedback was shown to improve teaching efficiency and robot performance within LfD. Focusing on robot performance in Figure~\ref{fig:performance}, the results imply that as users encounter difficulties, they can become more reluctant to continue teaching as performance plateaus or dips without explanation. On the other hand, participants in the EF condition were still able to complete the task despite drops in performance. Anecdotally, participant reluctance appears to largely stem from difficulties identifying areas which need instruction. For example, participant P16-NF in the no-feedback condition stated that they "saw no change [in the performance value], so I gave up", while P20-NF believed the performance value "didn't change much [and its] impact was minimal". On the other hand, when given explanatory feedback, P21-EF found that "the feedback helped me understand how to approach it, ... it’s not going to learn from one side". These responses highlight why teaching efficiency was improved. They demonstrate that explanatory feedback can guide users to which "side" or region requires teaching, allowing them to provide more efficient demonstrations and avoid or reduce performance plateaus. In contrast, users who did not receive feedback would often try to "iron out a path" (P28-NF) or realise after some time that "I could go to the green [non-preferred] goal" (P16-NF), resulting in inefficient demonstrations.  

\textbf{User Understanding}: This work also demonstrates that user prediction accuracy is improved by explanatory feedback within human-to-robot teaching. Looking at the results in Figure~\ref{fig:accuracy_bar}, we can see that prediction accuracy is improved by explanatory feedback. In response to explanatory feedback a participant stated "I know exactly what it was doing from the first feedback" (P25-EF), whilst another stated "Initially I knew nothing about how it learnt but eventually it looked like it was having trouble from the other side" (P17-EF). This suggests that explanatory feedback allowed participants to improve their understanding of the robot's behaviour either quickly or gradually. Furthermore, Table~\ref{tab:init_vs_fin_preds} demonstrates that prediction accuracy improved after teaching. Accuracy was greater in both \emph{final} prediction sub-tasks across both conditions when compared to the \emph{initial} prediction sub-tasks. Participants stated that their ability to predict "improved as [they were] able to recall how [they] trained it" (P22-NF), and that they understood "why the agent behaved that way, which is connected to the demonstrations [they] showed" (P19-EF). 
Additionally, the role of explanatory feedback might be most important during \emph{initial} user understanding, as this is when prediction accuracy is lowest and therefore where most misconceptions exist. This is supported by participant P19-EF who also stated that "The first [explanatory trajectories] kind of caught me off guard", demonstrating a difference between their mental-model and the robot's behaviour which required correction.
After experiencing the EF condition, participants stated that "the learning process only captured what I've shown" (P13-EF) and that "feedback let me know how well the robot learns my strategies, ... it lets me know which starting position doesn't fit it’s existing strategy" (P23-EF). This suggests that prediction accuracy and teaching efficiency are potentially linked and that improving user understanding can lead to more effective teaching. 

\textbf{User Perception and Certainty}: Surprisingly, though objective performance and user prediction accuracy were improved in the explanatory feedback condition, we did not see any significant differences in perceived understanding, performance, teaching, mental demand or prediction certainty between the two groups. A contributing factor to this is the between-participants study design. Since participants were not able to compare between the two conditions, it can be difficult to perceive the impacts of explanations or lack thereof. This is supported by user statements, where participants often responded positively to explanatory feedback whilst NF condition participants did state it was hard to gauge the impact of explanations despite the lack of significant difference. For example, P19-EF mentioned that "at first I wasn't quite sure about how everything was supposed to go, but after the first set of feedback, everything started to make sense and I understood what it should be doing" whereas P26-NF stated that "it's kind of hard to say [how their understanding changed] since I only saw [explanations] at the end". Perceived robot satisfaction and teaching could also be associated either with the \emph{number of demonstrations} and/or the \emph{final robot performance}, where a higher weighting on \emph{final robot performance} could lead to similar ratings when considering all but one participant achieved greater than 80\% performance eventually. Additionally, mental demand could be higher in the explanatory feedback condition, due to the uptake of information from feedback. Though there are no significant outcomes, Figure~\ref{fig:barplot} does show some deviations between the conditions, indicating that a larger population or a within-subjects study design may demonstrate some differences, motivating further investigation. 

Furthermore, though there was not an overall improvement in prediction certainty, the lower \emph{action prediction time} and significantly lower \emph{goal prediction time} paired with the improvement in overall prediction accuracy may suggest that explanatory feedback can reduce cognitive effort required to predict the robot's behaviour. We can additionally observe that certainty and its distribution are significantly higher and altered in the case of \emph{final goal prediction certainty}, where the prediction accuracy is highest (96\%). Conversely, we observe that action prediction certainties do not appear to change significantly between initial or final predictions as well as between conditions. This may be related to the nature of local actions, where multiple actions may be considered similarly optimal and therefore the most optimal action may be ambiguous. Another possible reason is the greater number of \emph{explanatory trajectories} viewed by EF condition participants. Viewing more trajectories may have allowed participants to gain a greater understanding of global behaviours over time as these are less likely to deviate between teaching sets when compared to local behaviours. This suggests that other explanatory feedback methods could be necessary to improve user certainty overall. Additionally, it indicates that trajectory-based explanatory feedback may be most suited to improving user certainty when their mental model closely aligns with the robot, as the only certainty metric which was significantly improved was the \emph{final goal prediction certainty} which had the highest prediction accuracy and lowest standard deviation.

\section{Conclusion and Future Work}
This work proposes a novel adaptive-sampling explanatory feedback system capable of generalizing to various LfD algorithms and adapting to user teaching. The system was evaluated through a user study which investigated its impact on performance, understanding, user perception and prediction certainty. The results demonstrated that explanatory feedback can improve human-to-robot teaching performance and understanding through improved robot performance, teaching efficiency and predictive ability as hypothesized. Importantly, user statements indicated that greater understanding leads to improved robot performance and teaching efficiency. However, despite these improvements, significant changes in overall user perception and prediction certainty were absent, although user statements indicated otherwise.

Further studies on explanatory feedback systems could benefit from different testing designs and explanatory feedback methodologies by better highlighting the impact of explanatory feedback, especially in comparison to a control. This can be done by exploring ways to disambiguate local robot behaviour as well as remove potential confounds. In our next work, we also intend to extend this work to a more complex robotic system and task. We will investigate how an adaptive explanatory feedback system can allow robot operators to make more effective choices. The overall goal is to develop a framework for explanatory feedback that can be generalised to a broad set of robotic and machine-learning systems.


\bibliographystyle{IEEEtran}
\bibliography{refs}

\begin{thebibliography}{10}
\providecommand{\url}[1]{#1}
\csname url@rmstyle\endcsname
\providecommand{\newblock}{\relax}
\providecommand{\bibinfo}[2]{#2}
\providecommand\BIBentrySTDinterwordspacing{\spaceskip=0pt\relax}
\providecommand\BIBentryALTinterwordstretchfactor{4}
\providecommand\BIBentryALTinterwordspacing{\spaceskip=\fontdimen2\font plus
\BIBentryALTinterwordstretchfactor\fontdimen3\font minus \fontdimen4\font\relax}
\providecommand\BIBforeignlanguage[2]{{%
\expandafter\ifx\csname l@#1\endcsname\relax
\typeout{** WARNING: IEEEtran.bst: No hyphenation pattern has been}%
\typeout{** loaded for the language `#1'. Using the pattern for}%
\typeout{** the default language instead.}%
\else
\language=\csname l@#1\endcsname
\fi
#2}}

\bibitem{Goel2020}
R.~Goel and P.~Gupta, ``Robotics and industry 4.0,'' \emph{Advances in Science, Technology and Innovation}, 2020.

\bibitem{niemela_telepresence_2021}
M.~Niemelä, L.~van Aerschot, A.~Tammela, I.~Aaltonen, and H.~Lammi, ``Towards ethical guidelines of using telepresence robots in residential care,'' \emph{International Journal of Social Robotics}, 2021.

\bibitem{service_robotics_2023_2030}
\BIBentryALTinterwordspacing
``Service robotics market size, share \& industry analysis,'' 2022. [Online]. Available: \url{https://www.fortunebusinessinsights.com/industry-reports/service-robotics-market-101805}
\BIBentrySTDinterwordspacing

\bibitem{ravichandar_lfd_2020}
H.~Ravichandar, A.~S. Polydoros, S.~Chernova, and A.~Billard, ``Recent advances in robot learning from demonstration,'' \emph{Annual Review of Control, Robotics, and Autonomous Systems}, 2020.

\bibitem{song_2020_wild}
S.~Song, A.~Zeng, J.~Lee, and T.~Funkhouser, ``Grasping in the wild: Learning 6dof closed-loop grasping from low-cost demonstrations,'' \emph{IEEE RA-L}, 2020.

\bibitem{ziebart_maxentirl_2008}
B.~D. Ziebart, A.~Maas, J.~A. Bagnell, and A.~K. Dey, ``Maximum entropy inverse reinforcement learning,'' in \emph{AAAI}, 2008.

\bibitem{Chen2003Programming}
J.~Chen and A.~Zelinsky, ``Programing by demonstration: Coping with suboptimal teaching actions,'' \emph{International Journal of Robotics Research}, 2003.

\bibitem{floden_2017_feedback}
J.~Flodén, ``The impact of student feedback on teaching in higher education,'' \emph{Assessment \& Evaluation in Higher Education}, 2017.

\bibitem{sena_quantifying_2020}
A.~Sena and M.~Howard, ``Quantifying teaching behavior in robot learning from demonstration,'' \emph{The International Journal of Robotics Research}, 2020.

\bibitem{Luebbers2021ARCLFD}
M.~B. Luebbers, C.~Brooks, C.~L. Mueller, D.~Szafir, and B.~Hayes, ``Arc-lfd: Using augmented reality for interactive long-term robot skill maintenance via constrained learning from demonstration,'' \emph{ICRA}, 2021.

\bibitem{adadi_peeking_2018}
A.~Adadi and M.~Berrada, ``Peeking {Inside} the {Black}-{Box}: {A} {Survey} on {Explainable} {Artificial} {Intelligence} ({XAI}),'' \emph{IEEE Access}, 2018.

\bibitem{Silva2023Subjective}
A.~Silva, M.~Schrum, E.~Hedlund-Botti, N.~Gopalan, and M.~Gombolay, ``Explainable artificial intelligence: Evaluating the objective and subjective impacts of xai on human-agent interaction,'' \emph{International Journal of Human–Computer Interaction}, 2023.

\bibitem{silva_personalized_2024}
A.~Silva, M.~Schrum, and M.~Gombolay, ``Towards balancing preference and performance through adaptive personalized explainability,'' \emph{ACM/IEEE International Conference on HRI}, 2024.

\bibitem{sakr2023everyday}
M.~Sakr, Z.~Zhang, B.~Li, H.~Zhang, H.~F. M.~V. der Loos, D.~Kulic, and E.~Croft, ``How can everyday users efficiently teach robots by demonstrations?'' \emph{arXiv preprint arXiv:2310.13083}, 2023.

\bibitem{barredo_arrieta_2020_XAI}
A.~B. Arrieta, N.~Díaz-Rodríguez, J.~D. Ser, A.~Bennetot, S.~Tabik, A.~Barbado, S.~Garcia, S.~Gil-Lopez, D.~Molina, R.~Benjamins, R.~Chatila, and F.~Herrera, ``Explainable artificial intelligence (xai): Concepts, taxonomies, opportunities and challenges toward responsible ai,'' \emph{Information Fusion}, 2020.

\bibitem{paleja_2021_utility}
R.~Paleja, M.~Ghuy, N.~Ranawaka~Arachchige, R.~Jensen, and M.~Gombolay, ``The utility of explainable ai in ad hoc human-machine teaming,'' in \emph{Advances in Neural Information Processing Systems}.\hskip 1em plus 0.5em minus 0.4em\relax Curran Associates, Inc., 2021.

\bibitem{gunning_XAIexplainable_2019}
D.~Gunning, M.~Stefik, J.~Choi, T.~Miller, S.~Stumpf, and G.-Z. Yang, ``\BIBforeignlanguage{en}{{XAI}—{Explainable} artificial intelligence},'' \emph{\BIBforeignlanguage{en}{Science Robotics}}, 2019.

\bibitem{Bach2015Pixelwise}
S.~Bach, A.~Binder, G.~Montavon, F.~Klauschen, K.~R. Müller, and W.~Samek, ``On pixel-wise explanations for non-linear classifier decisions by layer-wise relevance propagation,'' \emph{PLOS ONE}, 2015.

\bibitem{ribeiro-etal-2016-lime}
M.~Ribeiro, S.~Singh, and C.~Guestrin, ``{``}why should {I} trust you?{''}: Explaining the predictions of any classifier,'' in \emph{NAACL}, 2016.

\bibitem{Ribeiro_anchors_2018}
M.~T. Ribeiro, S.~Singh, and C.~Guestrin, ``Anchors: High-precision model-agnostic explanations,'' \emph{AAAI}, 2018.

\bibitem{qiang_graphlime_2022}
Q.~Huang, M.~Yamada, Y.~Tian, D.~Singh, and Y.~Chang, ``Graphlime: Local interpretable model explanations for graph neural networks,'' \emph{IEEE Transactions on Knowledge and Data Engineering}, 2022.

\bibitem{xu_2015_show_attend}
K.~Xu, J.~Ba, R.~Kiros, K.~Cho, A.~Courville, R.~Salakhudinov, R.~Zemel, and Y.~Bengio, ``Show, attend and tell: Neural image caption generation with visual attention,'' in \emph{ICML}, 2015.

\bibitem{das_explainable_2021}
D.~Das, S.~Banerjee, and S.~Chernova, ``Explainable ai for robot failures: Generating explanations that improve user assistance in fault recovery,'' in \emph{ACM/IEEE International Conference on HRI}, 2021.

\bibitem{adamson_why_2021}
T.~Adamson, D.~Ghose, S.~C. Yasuda, L.~J.~S. Shepard, M.~A. Lewkowicz, J.~Duan, and B.~Scassellati, ``Why {We} {Should} {Build} {Robots} {That} {Both} {Teach} and {Learn},'' in \emph{{ACM}/{IEEE} {International} {Conference} on HRI}, 2021.

\bibitem{Mueller2021Counterfactuals}
C.~Mueller, A.~Tabrez, and B.~Hayes, ``Interactive constrained learning from demonstration using visual robot behavior counterfactuals,'' \emph{Robotics: Science and Systems}, 2021.

\bibitem{Weitz2021Explain}
K.~Weitz, D.~Schiller, R.~Schlagowski, T.~Huber, and E.~André, ``“let me explain!”: exploring the potential of virtual agents in explainable ai interaction design,'' \emph{Journal on Multimodal User Interfaces}, 2021.

\end{thebibliography}
\end{document}